\documentclass[sigconf]{acmart}
\AtBeginDocument{%
  }

\setcopyright{acmlicensed}
\copyrightyear{2018}
\acmYear{2018}
\acmDOI{XXXXXXX.XXXXXXX}
\acmConference[Conference acronym 'XX]{Make sure to enter the correct
  conference title from your rights confirmation email}{June 03--05,
  2018}{Woodstock, NY}
\acmISBN{978-1-4503-XXXX-X/2018/06}

\usepackage[utf8]{inputenc} 
\usepackage[T1]{fontenc}    
\usepackage{hyperref}       
\usepackage{url}            
\usepackage{booktabs}       
\usepackage{siunitx} 
\usepackage{nicefrac}       
\usepackage{microtype}      
\usepackage{nicematrix} 
\usepackage{xcolor}         
\usepackage{amsmath}
\usepackage{mathtools}
\usepackage{amsthm}
\usepackage{graphicx}
\usepackage{wrapfig}
\usepackage{lipsum}
\usepackage{color, colortbl}
\usepackage{bbm}
\usepackage{subcaption}
\usepackage{multirow}
\usepackage{tikz}
\usepackage{enumitem}
\usepackage{xspace}
\usepackage[most]{tcolorbox}
\tcbuselibrary{listingsutf8}
\usepackage{makecell}
\usepackage{arydshln}
\newcolumntype{g}{>{\cellcolor{Gray}}c}
\usepackage{subcaption}

\definecolor{citecol}{HTML}{2E8B57}
\definecolor{tableofcontent}{HTML}{E63E15}
\definecolor{urlcol}{HTML}{2470D8}
\usepackage[normalem]{ulem}
\useunder{\uline}{\ul}{}
\hypersetup{
    colorlinks=true,       
    linkcolor=tableofcontent, 
     citecolor=citecol,        
    urlcolor=black,           
}

\newcommand{\pgftextcircled}[1]{
    \setbox0=\hbox{#1}%
    \dimen0\wd0%
    \divide\dimen0 by 2%
    \begin{tikzpicture}[baseline=(a.base)]%
        \useasboundingbox (-\the\dimen0,0pt) rectangle (\the\dimen0,1pt);
        \node[circle,draw,outer sep=0pt,inner sep=0.1ex] (a) {#1};
    \end{tikzpicture}
}
\setlist{leftmargin=10mm}
\definecolor{Gray}{gray}{0.9}
\newcommand{\xhdr}[1]{{\vspace{1pt}\noindent\bfseries #1}.}
\newcommand{\ie}{\textit{i.e., }}
\newcommand{\eg}{\textit{e.g., }}

\newcommand{\wrt}{\textit{w.r.t. }}

\usepackage[most]{tcolorbox}
\usepackage{lipsum}
\tcbuselibrary{skins}
\usepackage{mdframed}
\definecolor{niceblue}{HTML}{3c9dfd}
\usepackage{fontawesome5}

\newtcolorbox{Mycolorbox}[2][]{
  arc=5mm,
  lower separated=false,
  fonttitle=\bfseries,
  colbacktitle=gray,
  coltitle=white,
  enhanced,
  attach boxed title to top left={xshift=0.5cm, yshift=-2mm},
  colframe=gray,
  colback=white,
  title=#1, #2,
  breakable
}

\usepackage[capitalize,noabbrev]{cleveref}
\usepackage{etoolbox}
\AtBeginEnvironment{lemma}{\hspace*{2em}}
\newtheorem{lemma}{Lemma} 
\theoremstyle{plain}
\newtheorem{theorem}{Theorem}[section]
\newtheorem{proposition}[theorem]{Proposition}

\theoremstyle{definition}

\theoremstyle{remark}

\copyrightyear{2026}
\acmYear{2026}
\setcopyright{cc}
\setcctype{by}
\acmConference[WWW '26]{Proceedings of the ACM Web Conference 2026}{April 13--17, 2026}{Dubai, United Arab Emirates}
\acmBooktitle{Proceedings of the ACM Web Conference 2026 (WWW '26), April 13--17, 2026, Dubai, United Arab Emirates}
\acmPrice{}
\acmDOI{10.1145/3774904.3792656}
\acmISBN{979-8-4007-2307-0/2026/04}

\begin{document}

\title{Towards A Universal Graph Structural Encoder}

\author{Jialin Chen}
\affiliation{
  \department{Department of Computer Science}
  \institution{Yale University}
  \city{New Haven}
  \state{CT}
  \country{United States}
}
\email{jialin.chen@yale.edu}

\author{Haolan Zuo}
\affiliation{
  \department{Department of Computer Science}
  \institution{Yale University}
  \city{New Haven}
  \state{CT}
  \country{United States}
}
\email{haolanzuo@gmail.com}

\author{Haoyu Wang}
\affiliation{
  \department{Department of Electrical and Computer Engineering}
  \institution{Georgia Institute of Technology}
  \city{Atlanta}
  \state{GA}
  \country{United States}
}
\email{haoyu.wang@gatech.edu}

\author{Siqi Miao}
\affiliation{
  \department{Department of Electrical and Computer Engineering}
  \institution{Georgia Institute of Technology}
  \city{Atlanta}
  \state{GA}
  \country{United States}
}
\email{siqi.miao@gatech.edu}

\author{Pan Li}
\affiliation{
  \department{Department of Electrical and Computer Engineering}
  \institution{Georgia Institute of Technology}
  \city{Atlanta}
  \state{GA}
  \country{United States}
}
\email{panli@gatech.edu}

\author{Rex Ying}
\affiliation{
  \department{Department of Computer Science}
  \institution{Yale University}
  \city{New Haven}
  \state{CT}
  \country{United States}
}
\email{rex.ying@yale.edu}

\renewcommand{\shortauthors}{Jialin Chen et al.}





\begin{abstract}
Recent advancements in large-scale pre-training have shown the potential to learn generalizable representations for downstream tasks. In the graph domain, however, capturing and transferring structural information across different graph domains remains challenging, primarily due to the inherent differences in graph topological patterns across various contexts. For example, a social network's structure is fundamentally different from that of a product co-purchase graph. Additionally, most existing models struggle to capture the rich topological complexity of graph structures, leading to inadequate exploration of the graph embedding space. To address these challenges, we propose GFSE, a universal pre-trained graph encoder designed to capture transferable structural patterns across diverse domains such as the web graph, social networks, and citation networks. GFSE is the first cross-domain graph structural encoder pre-trained with multiple self-supervised learning objectives. Built on a Graph Transformer, GFSE incorporates attention mechanisms informed by graph structural information, enabling it to encode intricate multi-level and fine-grained topological features within complex graph structures. The pre-trained GFSE produces generic and theoretically expressive positional and structural encoding for graphs, which can be seamlessly integrated with various downstream graph feature encoders, including graph neural networks for vectorized features and Large Language Models (LLMs) for text-attributed graphs. Comprehensive experiments on synthetic and real-world datasets demonstrate GFSE's capability to significantly enhance the model's performance while requiring substantially less task-specific fine-tuning. Notably, GFSE achieves state-of-the-art performance in diverse evaluated cases spanning various graph models and datasets, highlighting its potential as a powerful and versatile encoder for graph-structured data.
\end{abstract}

\maketitle

\section{Introduction}
The success of large-scale pre-trained models in domains such as natural language processing~\cite{achiam2023gpt, bubeck2023sparks, touvron2023llama}, computer vision~\cite{radford2021learning, ramesh2021zero}, audio~\cite{yang2023uniaudio, borsos2023audiolm} has demonstrated the power of learning transferable representations across diverse datasets. 
However, developing such foundational models for the graph domain remains a significant challenge. Due to the inherent heterogeneity of graph-structured data across the domains, most prior graph pre-training efforts are highly specialized, focusing on narrow domains such as link structures in web graphs, user interactions in social networks, or relational facts in knowledge graphs~\cite{galkin2023towards, zhang2025litfm, xia2024anygraph}. These specialized models often rely heavily on domain-specific features, which limits their transferability and requires deep domain expertise to adapt.

This feature-centric approach overlooks a critical, unifying aspect of graphs: they exhibit rich structural patterns that are often shared across domains. For example, social networks and citation networks commonly exhibit small-world properties and dense community structures, while e-commerce graphs reveal hierarchical taxonomies. Despite the prevalence of these fundamental patterns, they are often secondary to domain-specific features, hindering the generalization capabilities of existing pre-trained models.

As tokens in NLP derive meaning from their context, nodes in graphs gain significance through their structural role. Therefore, we propose a paradigm shift that prioritizes learning universal structural attributes—domain-agnostic features that are shared across diverse graph domains. These attributes, such as motifs, node degrees, and hierarchical topologies, serve as the building blocks for transferable graph representations. The central challenge lies in developing a foundational model that can effectively capture these universal structural patterns while remaining adaptable for downstream tasks. Such a model would reduce the reliance on domain-specific engineering and unlock the potential for truly general-purpose graph representation learning.

\begin{figure*}[t]\vspace{-0.3cm}
    \centering
    \includegraphics[width=0.9\textwidth]{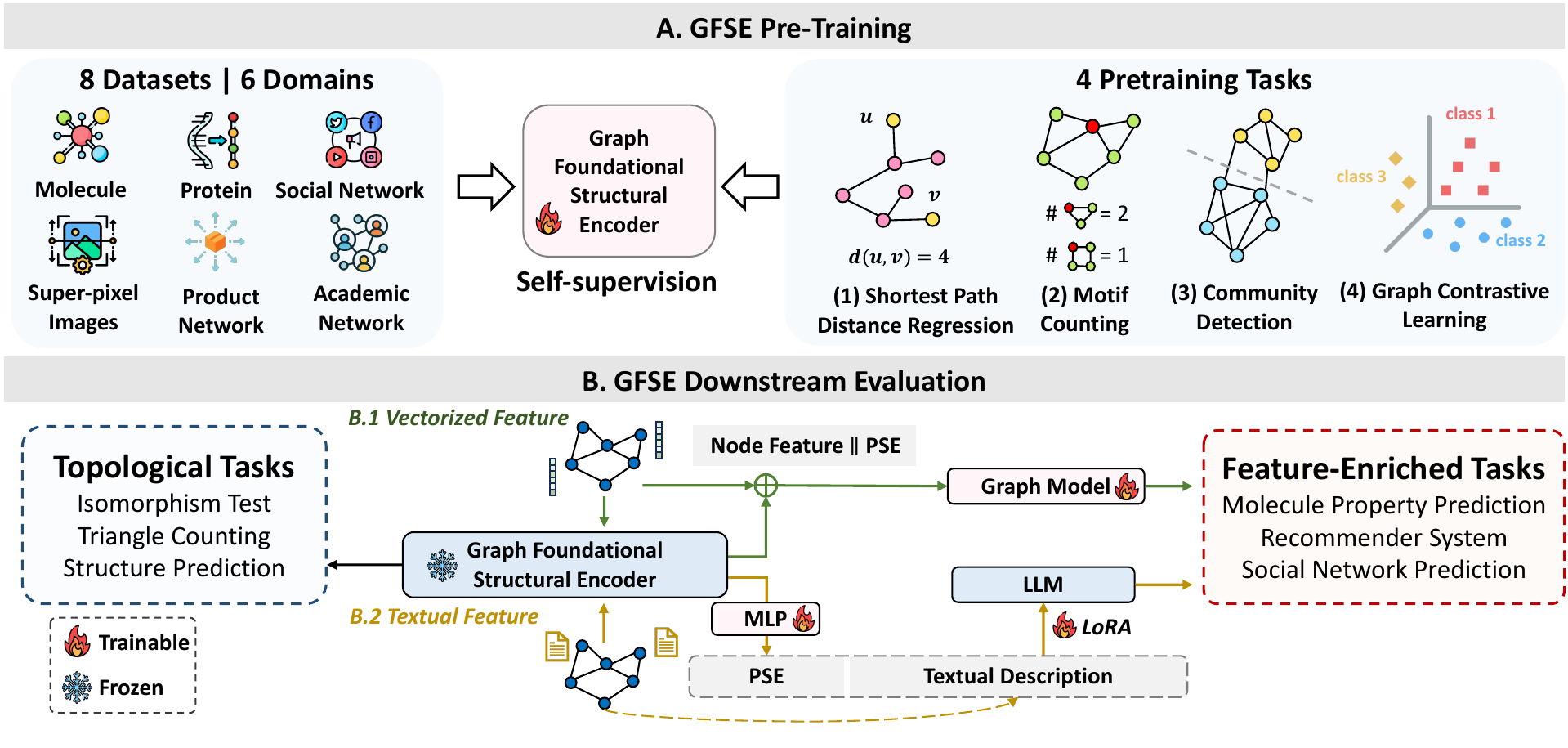}\vspace{-0.4cm}
    \caption{A) GFSE is pre-trained on 8 datasets from 6 different domains. Pre-training tasks include reconstruction and contrastive learning. B) GFSE generates generic and expressive Positional and Structural Encoding (PSE) to tackle topological graph tasks, which can also be seamlessly integrated into downstream feature encoders for feature-enriched tasks by concatenating with initial vectorized features or prepending the generated PSE to the textual prompt as a soft token.} 
    \label{fig:pipeline}\vspace{-0.4cm}
\end{figure*}

\xhdr{Proposed work} To address the challenges of cross-domain pre-training and effectively capturing universal structural encoding, we propose \textbf{GFSE}, a \textbf{G}raph \textbf{F}oundational \textbf{S}tructural \textbf{E}ncoder, as shown in Figure~\ref{fig:pipeline}. GFSE is pre-trained across diverse graph domains using multiple self-supervised pre-training tasks, including shortest path distance regression, motif counting, local community detection, and graph-level contrastive learning. Each pre-training task targets a critical and necessary aspect of graph structures, enabling GFSE to capture a comprehensive understanding of graph topology. GFSE employs a Graph Transformer enhanced with biased attention mechanisms. Notably, the relative positional encoding, derived from the random walk matrix, is explicitly integrated into the attention bias term. This design allows GFSE to effectively capture intricate structural dependencies among node pairs during pre-training, ensuring both efficiency and theoretically guaranteed expressiveness. GFSE's versatility extends to various graph learning scenarios. The pre-trained GFSE can produce generic and expressive Positional and Structural Encodings (PSE) for topological tasks. In feature-enriched contexts, the generated PSE can seamlessly augment vectorized features or integrate with language models for text-attributed graphs, enabling GFSE to serve as a powerful component in graph foundation models.

The \textbf{contributions} of this work are threefold. (1) We propose GFSE, a unified cross-domain graph structural encoder pre-trained with four essential self-supervised learning objectives. Extensive experiments demonstrate the effectiveness of these pre-training tasks, resulting in state-of-the-art performance across diverse evaluated cases, encompassing various graph models and datasets. (2) We provide theoretical justification and empirical results demonstrating GFSE's ability to generate expressive PSE. (3) GFSE serves as a plug-and-play solution for any graph model to incorporate generalizable structural information, reducing the need for domain-specific fine-tuning on heterogeneous web-scale tasks.


\section{Related Work}
\definecolor{hy}{RGB}{34, 139, 34}
\newcommand{\hy}[1]{{\color{hy}#1}}

\xhdr{Graph Pre-training}
Graph self-supervised learning approaches are typically pre-training graph models, \eg GNNs or Graph Transformers, on a massive amount of labeled graphs with inherent features by reconstructing the structures or masked attributes~\cite{cui2020adaptive, hou2022graphmae, kipf2016variational, hu2020gpt, wang2017mgae, xia2024opengraph, xia2024anygraph, zhao2024graphany, mizera2024graph}. Some works also utilize contrastive learning to enhance node and graph-level representation learning~\cite{han2022g,hassani2020contrastive,velickovic2019deep,hu2019strategies, lee2022augmentation, li2021pairwise, lu2021learning, sun2019infograph, sun2021mocl, xu2021infogcl, galkin2023towards, zhao2024all}. These methods, while effective in certain domains, exhibit limited generalizability across different graph domains due to their tailored design for specific types of data. Additionally, there have been graph prompt techniques~\cite{huang2024prodigy, fang2024universal} that can be used to enhance model adaptation over graphs. There have also been some attempts at cross-domain graph pre-training models~\cite{qiu2020gcc,davies2023its}. Unfortunately, all these models rely on one singular pre-training task (\ie contrastive learning), and usually fail to capture fine-grained structural features at the node level or edge level~\cite{mao2024graph}. 

\xhdr{Generalizable Positional and Structural Encoding (PSE)} Early research on the web graph focused on characterizing its macroscopic properties, such as power-law degree distributions and the prevalent 'bow-tie' structure~\cite{broder2000graph, faloutsos1999power, albert1999internet}, which led to seminal node-importance algorithms like PageRank~\cite{brin1998anatomy}. This perspective inspired a family of traditional, hand-crafted Positional and Structural Encodings (PSEs) based on such structural statistics, including Laplacian PE~\cite{davies2024generalised, kreuzer2021rethinking, beaini2021directional, wang2022equivariant}, shortest-path distance~\cite{li2020distance, ying2021transformers}, random-walk encodings~\cite{ma2023graph, dwivedi2021graph, bruel2022rewiring, rampavsek2022recipe}, and node degree~\cite{ying2021transformers}. However, these global metrics and engineered features often fail to capture the complex, local connectivity patterns and subgraph motifs that signify sophisticated behaviors like spam link farms or coordinated user activity~\cite{milo2002network, leskovec2007graph}. To overcome these limitations, recent work has shifted towards models that learn PSE adaptively~\cite{kreuzer2021rethinking, dwivedi2021graph, chen2022structure, lim2022sign}. While some methods like GPSE~\cite{liu2023graph} pre-train a structural encoder, they still suffer from limited cross-domain transferability due to simplistic backbones. Developing an effective and versatile PSE remains an open challenge that requires further innovation.



\xhdr{Graph Language Models}
With the success of foundation models in NLP, recent efforts also harness Large Language Models (LLMs) to develop domain-specific graph foundation models. This typically involves converting graph data into a text-compatible format by linearizing the graph's structure (\eg as a sequence of adjacency lists or edge sets) and verbalizing node/edge attributes to create descriptive prompts~\cite{chen2024llaga, tang2023graphgpt,ye2023natural,qian2023can, zhao2023graphtext,guo2023gpt4graph,chen2024exploring, liu2023one,kong2024gofa, chen2024text, fan2024graph, zhang2024graphtranslator,li2024graph}. The goal is to create a single, powerful model that can reason over both the semantics of text attributes and the graph's topology. However, this approach faces a fundamental challenge: LLMs, which are pre-trained on sequential text, struggle to comprehend the complex, non-Euclidean nature of graph structures. Recent empirical studies and benchmarks consistently show that serializing a graph into a one-dimensional sequence leads to a significant loss of topological information, impairing an LLM's ability to reason about complex connectivity, distances, and higher-order patterns~\cite{fatemi2023talk, wang2024can, wang2024nlgraph}. Our work addresses this gap directly. Instead of relying on a single model for both text and structure, we focus on developing an expressive graph model dedicated solely to encoding rich topological information. Our approach is designed to be complementary to LLMs, serving as a universal structural encoder that can provide the crucial topological context that LLMs inherently lack, thereby enabling a more powerful and synergistic approach for text-attributed graphs.

\section{Proposed Method}
As shown in Figure~\ref{fig:pipeline}, we collect graph pre-training datasets from six different domains, including social networks, product networks, academic networks, images, molecules, and proteins. GFSE utilizes a transformer-based architecture with biased attention to incorporate relative inductive bias within graph structures (Sec.~\ref{sec:archi}). GFSE is pre-trained with four challenging self-supervision tasks simultaneously, each designed to enhance a crucial aspect of structural awareness and promote encoding quality (Sec.~\ref{sec:pretraining_task}).  GFSE generates expressive positional and structural encoding (PSE) for topological tasks (Sec.~\ref{sec:expressive}). The generated PSE can also be seamlessly integrated into graphs with vectorized features or textual features, enhancing the downstream performance (Sec.~\ref{sec:downstream_method}).
\subsection{Architecture}\label{sec:archi}
Previous work~\cite{liu2023graph} uses randomized features to replace initial node features. However, it leads to poor generalizability across different domains. In this work, we propose to use both absolute and relative random-walk positional encoding as the initial features. Formally, let $G(V,E)$ represent an input graph, where $V$ and $E$ denote the set of nodes and edges, respectively. $\mathbf{A}\in\mathbb{R}^{N\times N}$ indicates the adjacency matrix, where $N$ is the number of nodes, and $\mathbf{D}$ is the degree matrix. Random Walk matrix is defined as $\mathbf{M} = \mathbf{D}^{-1}\mathbf{A}$, where $\mathbf{M}_{i,j}$ indicates the transition probability from the $i$-th node to the $j$-th node. Following previous works~\cite{ma2023graph}, we calculate the $d$-dimensional encoding for each node and node pairs as follows.
\begin{equation}\label{eq:rw}
    \mathbf{P}_i= [\mathbf{I}, \mathbf{M}, \mathbf{M}^2, \cdots,\mathbf{M}^{d-1}]_{i,i}, \quad 
    \mathbf{R}_{i,j}= [\mathbf{I}, \mathbf{M}, \mathbf{M}^2, \cdots,\mathbf{M}^{d-1}]_{i,j}
\end{equation}
$\mathbf{P}\in\mathbb{R}^{N\times d}$ and $\mathbf{R}\in\mathbb{R}^{N\times N\times d}$ are used as the initial node features and edge features in GFSE.

\xhdr{Pre-training Backbone} GFSE is built on a GPS architecture~\cite{rampavsek2022recipe} for pre-training due to its scalability and generalizability. Each GPS layer contains local message passing and global attention modules to capture neighbor and long-range information. In the $\ell$-th layer, the node encoding $\mathbf{P}^{\ell}$ and relative edge encoding $\mathbf{R}^{\ell}$ are fed into message passing layers ($\operatorname{MPNN}$) and Biased Attention Module ($\operatorname{BiasAttn}$) parallelly. The node encoding is updated by aggregating local and global information $\mathbf{P}^{\ell+1} =\operatorname{MLP}^{\ell}\left(\mathbf{P}_M^{\ell+1}+\mathbf{P}_T^{\ell+1}\right)$ where
\begin{equation}\label{eq:attn_bias}
    \mathbf{P}_M^{\ell+1}, \mathbf{R}^{\ell+1} =\operatorname{MPNN}^{\ell}\left(\mathbf{P}^{\ell}, \mathbf{R}^{\ell}, \mathbf{A}\right); \quad
\mathbf{P}_T^{\ell+1} =\operatorname{BiasAttn}^{\ell}\left(\mathbf{P}^{\ell}, \mathbf{R}^{\ell}\right).
\end{equation}
\xhdr{Attention Bias} The global attention in the original GPS framework does not account for relative edge encoding while leaving them entirely for the message-passing layers. However, incorporating relative edge encoding in global attention is crucial for capturing long-range dependencies, as the receptive field of message-passing layers is inherently constrained by their depth. 
To build a theoretically more powerful encoder, we explicitly incorporate relative edge encoding into global attention, where the attention weight between the $i$-th and the $j$-th nodes is computed by $a_{i,j}^\prime = \operatorname{SoftMax}(a_{i,j} + \operatorname{Linear}(\mathbf{R}_{i,j}))$, where $\operatorname{Linear}: \mathbb{R}^{d}\rightarrow \mathbb{R}$ indicates a linear layer that maps the $d$-dimensional relative encoding to a scalar. $a_{i,j}$ denotes the original attention weight computed by scaled-dot self-attention on the node encoding $\mathbf{P}^{\ell}$ in each GPS layer. 


\subsection{Expressive Power of GFSE}\label{sec:expressive}

We show that GFSE can generate highly expressive PSE by incorporating relative edge encoding into the attention computation. To characterize the expressiveness of GFSE, we employ the Structural Encoding enhanced Global Weisfeiler-Lehman test (SEG-WL)~\cite{zhu2023structural}, a generalized WL test that incorporates relative structural encoding into the isomorphism algorithm. For an input graph $G(V,E)$ with node set $V$ and edge set $E$, let $f_P:V\rightarrow \mathcal{X}$ and $f_R:V\times V\rightarrow\mathcal{X}$ indicate the node-level and edge-level structural encoding, respectively. Different from traditional WL test, SEG-WL updates the node labels at the $t$-th iteration by $g_t(v)=\text{hash}\left(\left\{\left\{\left(g_{t-1}(u), f_R(v, u)\right): u \in V\right\}\right\}\right)$ and $g_0(v)=\text{hash}(f_P(v))$. SEG-WL can be viewed as a high-level abstraction of the learning paradigm of our pre-training architecture with biased attention (Eq.~\ref{eq:attn_bias}), where relative structural encoding between any two nodes is considered for updating node representations~\cite{zhu2023structural}. Let RW($d$)-SEG-WL denote the case that $f_P$ and $f_R$ are determined by $\mathbf{P}$ and $\mathbf{R}$ with $d$ dimension, \ie $f_P(v_i)=\mathbf{P}_i\in\mathbb{R}^d$ and $f_R(v_i, v_j)=\mathbf{R}_{ij}\in\mathbb{R}^d$ for the $i$-th and $j$-th nodes. We have the following propositions.
\begin{proposition}\label{prop1}
    RW($d$)-SEG-WL ($d\geq 3$) is strictly more expressive than 1-WL in testing non-isomorphic graphs.
\end{proposition} \vspace{-0.3cm}
\begin{proof}
We first introduce Neighbor-SEG-WL, which is the SEG-WL test when $f_P$ is an identity encoding and $f_R(u,v)$ equals 1 if $(u,v)\in E$ and $2$ otherwise. Previous works have proved the following Proposition~\cite{zhu2023structural}:
\begin{lemma}
    Two non-isomorphic graphs can be distinguished by WL if and only if they are distinguishable by Neighbor-SEG-WL.
\end{lemma}
Therefore, Neighbor-SEG-WL is a specific example of SEG-WL test that has equivalent expressiveness to the 1-WL test. We then prove that RW-SEG-WL is strictly more expressive than Neighbor-SEG-WL. Let $d_{\text{neg}}(u,v)$ indicate the edge-level encoding $f_R$ in Neighbor-SEG-WL. Note $d_{\text{neg}}(v_i,v_j)=2$ if and only if $A_{ij}=0$. Recall that $f_R(\cdot,\cdot)$ in RW-SEG-WL satisfies $f_R(v_i, v_j)=\mathbf{R}_{ij}\in\mathbb{R}^d$ with $\mathbf{R}=[\mathbf{I},\mathbf{M},\cdots, \mathbf{M}^{d-1}]$ where $\mathbf{M}=\mathbf{D}^{-1}\mathbf{A}$. Therefore, $f_R$ in RW-SEG-WL strictly contains the information of $d_{\text{neg}}$. Therefore, if two non-isomorphic graphs can be distinguished by WL, they can be distinguished by RW-SEG-WL. Proposition~\ref{prop1} is proved. 
\end{proof}
\begin{figure}[h]\vspace{-0.3cm}
\centering
\includegraphics[width=0.3\textwidth]{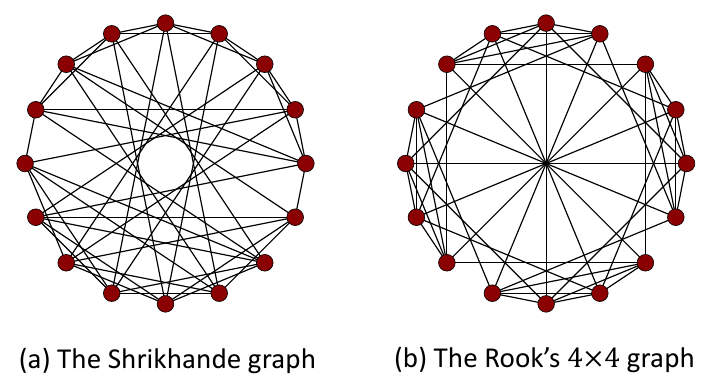} \vspace{-0.3cm}
\caption{RW($d$)-SEG-WL can distinguish two non-isomorphic graphs when $d>4$ while 3-WL fails.}\vspace{-0.3cm}
\label{fig:iso_graph}
\end{figure}
\begin{proposition}\label{prop2}
    There exist pairs of graphs that RW($d$)-SEG-WL can distinguish, but 3-WL can not.
\end{proposition}\vspace{-0.3cm}
\begin{proof}
To prove Proposition~\ref{prop2}, we provide an example in Figure~\ref{fig:iso_graph} which shows the Shrikhande graph and the Rook's $4\times 4$ graph, a pair of strongly regular graphs. It has been proved that they cannot be distinguished by 3-WL~\cite{arvind2020weisfeiler}. We empirically verified that RW($d$)-SEG-WL with $d>4$ can distinguish these two graphs.
\end{proof}\vspace{-0.2cm}
\begin{table*}[htbp]
\caption{Results of synthetic graph isomorphism tests} \label{tab:synthetic}
\centering\vspace{-0.3cm}
\resizebox{0.9\textwidth}{!}{
\begin{tabular}{lcccccccccc}
\toprule
 & \multicolumn{4}{c}{\textbf{Low-Order Graphs (Parameter:$n$)}} & \multicolumn{6}{c}{\textbf{Strongly Regular Graphs (Parameter:($n,k,\lambda,\mu$)}} \\ \midrule
 Parameter & 5 & 6 & 7 & 8 & \multicolumn{1}{l}{(25,12,5,6)} & \multicolumn{1}{l}{(26,10,3,4)} & \multicolumn{1}{l}{(29,14,6,7)} & \multicolumn{1}{l}{(36,14,4,6)} & \multicolumn{1}{l}{(40,12,2,4)} & \multicolumn{1}{l}{45,12,3,3)} \\ \midrule
\# graphs & 21 & 112 & 853 & 11117 & 15 & 10 & 41 & 180 & 28 & 78 \\
\# graph pairs & 210 & 6216 & 363378 & 61788286 & 105 & 45 & 820 & 16110 & 378 & 3003 \\ \midrule
 \multicolumn{11}{c}{\textbf{number of undistinguishable graph pairs}} \\ \midrule
WL & 0 & 3 & 17 & 312 & 105 & 45 & 820 & 16110 & 378 & 3003 \\
SPD-SEG-WL & 0 & 2 & 12 & 186 & 105 & 45 & 820 & 16110 & 378 & 3003 \\
\rowcolor{Gray}\textbf{RW-SEG-WL($d=8$) }& \textbf{0} &\textbf{0}&\textbf{0}&\textbf{0}&\textbf{0}&\textbf{0}&\textbf{0}&\textbf{0}&\textbf{0}&\textbf{0} \\
\bottomrule
\end{tabular}}\vspace{-0.3cm}
\end{table*}

\xhdr{Synthetic Graph Isomorphism Tests} To evaluate the expressive power of RW-SEG-WL, we perform synthetic graph isomorphism tests on low-order graphs and strongly regular graphs. We consider low-order graphs with up to 8 nodes. Strongly regular graph pair SRG($n,k,\lambda,\mu)$ means graphs with $n$ nodes, where each node has $k$ neighbors. Each adjacent pair of nodes has the same number $\lambda$ of neighbors in common, and each non-adjacent node pair has $\mu$ neighbors in common. Strongly regular graphs are known to be challenging cases for graph isomorphism test algorithms due to their highly symmetric structure~\cite{arvind2020weisfeiler}. We compare with 1-WL and SPD-SEG-WL, where $f_P$ is an identity encoding and $f_R$ is defined as the shortest path distance between two nodes. Note SPE-SEG-WL can be viewed as an expressivity upper bound of Graphormer~\cite{ying2021transformers}. The results are shown in Table~\ref{tab:synthetic}. We observe that RW-SEG-WL can distinguish significantly more non-isomorphic graphs than 1-WL and SPD-SEG-WL. Specifically, with $d$ equals 8, \ie when considering a random walk with 8 steps, RW-SEG-WL successfully distinguishes all low-order graphs and strongly regular graphs. When setting $d=4$, there are 16 pairs of strongly regular graphs that cannot be distinguished by RW-SEG-WL. Therefore, it is natural to develop a graph transformer equipped with a relative random-walk encoding that can accurately capture important graph structures and demonstrate strong expressive power.

RW-SEG-WL stands as an expressivity upper bound of our proposed GFSE. The pre-training tasks are meticulously designed to push GFSE towards achieving the upper bound established by RW-SEG-WL. The pre-training tasks are carefully designed to optimize both node-level and edge-level structural encoding, progressively refining the effectiveness of the model in generating expressive encoding.


\subsection{Self-supervised Pre-training Tasks}\label{sec:pretraining_task}
GFSE is pre-trained with reconstruction and contrastive structural tasks at multiple context levels. Each task highlights a specific structural aspect, thereby augmenting the model's expressiveness and capability to capture complex graph structures. Let $\mathbf{P}^L\in\mathbb{R}^{N\times d_e}$ represent the output after $L$ layers. We decode $\mathbf{P}^L$ with independent MLP heads for each pre-training task.

\textbf{Shortest Path Distance Regression} encodes the global proximity and connectivity between nodes, which helps to discern nodes' positions and relations within the entire graph~\cite{li2020distance}. We pre-compute the shortest path distance via the Dijkstra algorithm~\cite{dijkstra1959note} to create the label $\text{SPD}\in\mathbb{R}^{N\times N}$. The loss for shortest path distance regression is computed by $\mathcal{L}_{\text{SPD}} =\frac{1}{|E|}\sum_{i,j\in V}\|h_{\text{SPD}}(\mathbf{P}^L_i\|\mathbf{P}^L_j)-\text{SPD}_{i,j}\|^2$,
where $\|$ indicates the concatenation and $h_{\text{SPD}}$ indicates a task-specific head.

\textbf{Motif Counting} enables the model to better identify each node's role in the surrounding subgraphs.  To improve expressiveness, we include a variety of small motifs, called \textit{graphlets}~\cite{prvzulj2004modeling, prvzulj2007biological, bouritsas2022improving}, with different numbers of vertices, which are beyond usual types like stars, paths, cycles, and cliques. Let $Y_i^m\in\mathbb{Z}^k$ denote the motif label that indicates the number of certain motifs surrounding node $i$, where $k$ is the number of graphlet types. The loss is formulated as $\mathcal{L}_{\text{MC}} = \frac{1}{|V|}\sum_{i\in V}\|h_{\text{MC}}(\mathbf{P}^L_i)-Y_i^m\|^2$, where $h_{\text{MC}}:\mathbb{R}^d_e\rightarrow\mathbb{R}^k$ is the task-specific head for motif counting. 

\textbf{Community Detection} aims to identify densely connected subgraphs, where nodes within a community are more closely linked to each other than to nodes outside the community. Such community structures are ubiquitous in various real-world networks, \eg social networks, and transportation systems. We employ the Louvain Community Detection Algorithm~\cite{blondel2008fast} to extract the community structure from pre-training graphs, which clusters nodes into communities based solely on graph topology without node features. We approach this task in a contrastive learning manner by minimizing the embedding distances between intra-community nodes while maximizing the distance between inter-community nodes.
\vspace{-0.2cm}
\begin{equation*}
        \mathcal{L}_{\text{CD}} = \sum_{i\in V}\sum_{j\in V} Y_{i,j}^c (1-\operatorname{sim}(i,j) )
        + (1-Y_{i,j}^c)\max(0, \epsilon - (1- \operatorname{sim}(i,j))), \vspace{-0.2cm}
\end{equation*}
where the similarity score $\operatorname{sim}(i,j)$ is calculated by $\operatorname{sim}(i,j) = \frac{\boldsymbol{z}_i \cdot \boldsymbol{z}_j}{||\boldsymbol{z}_i||\cdot || \boldsymbol{z}_j||}$ and $\boldsymbol{z}_i=h_{\text{CD}}(\mathbf{P}^L_i)$ with a head $h_{\text{CD}}$. $\epsilon$ is a margin hyperparameter. $Y_{i,j}^c$ is a binary label that indicates if the $i$-th node and the $j$-th node are in the same community. Through $\mathcal{L}_{\text{CD}}$, GFSE learns to discern community boundaries and distinguish nodes based on local community structures.

\textbf{Graph Contrastive Learning} aims to distinguish graphs from different datasets. The motivation is rooted in the observation that similar structural characteristics in different domains may exhibit distinct meanings. For example, a densely connected subgraph in a citation network might correspond to a specific research topic, whereas a similar structure in a social network could represent a tightly-knit community or interest group. Therefore, GFSE distinguishes graphs from different datasets with the contrastive loss $\mathcal{L}_{\text{GCL}} =-\log \frac{\exp \left(\operatorname{sim}\left(\boldsymbol{z}_{G_i}, \boldsymbol{z}_{G_j}\right) / \tau\right)}{\sum_{k=1}^{K} \mathbbm{1}_{[G_k \nsim G_i]} \exp \left(\operatorname{sim}\left(\boldsymbol{z}_{G_i}, \boldsymbol{z}_{G_k}\right) / \tau\right)}$, where $\tau$ is the temperature, $G_i$ and $G_j$ are from the same dataset, $\boldsymbol{z}_{G_i}=\mathrm{GlobalPool}(h_{\text{GCL}}(\mathbf{P}^L_{G_i}))$ is the output of the global pooling applied to the last layer's representation $\mathbf{P}^L_{G_i}$ for the graph $G_i$, $K$ is the number of negative samples, and $\mathbbm{1}_{[G_k \nsim G_i]}$ indicates whether graphs $G_k$ and $G_i$ originate from different datasets. 

\textbf{Multi-task Selection and Loss Balance:} Each pre-training task targets a different structural aspect, enabling GFSE to capture a comprehensive understanding of graph topology. For instance, the shortest path distance regression task emphasizes global connectivity within graphs, while motif counting focuses on the occurrence of local subgraph patterns. Although other pre-training tasks could be considered, the chosen tasks provide a balanced and effective combination, capable of capturing both local and global structural information at varying granularities without introducing unnecessary complexity. Since the loss scale and difficulty vary significantly across tasks, we employ task-specific uncertainty~\cite{kendall2018multi} to unify the loss scales, which automatically balances different pre-training losses, \ie $\mathcal{L}_{\text{SPD}}, \mathcal{L}_{\text{MC}}, \mathcal{L}_{\text{CD}}$, and $ \mathcal{L}_{\text{GCL}}$ (see Appendix~\ref{app:uncertainty} for more details). Moreover, tracking the evolution of uncertainty values offers an interpretable perspective on how each task contributes to the overall pre-training process, enabling a deeper understanding of the model’s learning dynamics and task interactions.

\subsection{Combination with Downstream Feature Encoder}\label{sec:downstream_method}
\xhdr{Application on Graphs with Vectorized Features} GFSE can be readily employed to generate expressive PSE for various graph applications. Let $\mathbf{X}^0 \in \mathbb{R}^{N \times d_x}$ denote the initial node features for a given graph with $N$ nodes and $\mathbf{P}^L \in \mathbb{R}^{N \times d_e}$ denote PSE generated by GFSE, where $d_x$ and $d_e$ are dimensions of node features and PSE, respectively. $\mathbf{P}^L$ can then be concatenated with the initial node features $\mathbf{X}^0$ to create a new feature matrix $\mathbf{X}^{new} = [\mathbf{X}^0 \| \mathbf{P}^L] \in \mathbb{R}^{N \times (d_x + d_e)}$, which augments the node features with structural information. This structure-enriched feature $\mathbf{X}^{new}$ can subsequently be fed into downstream graph models, such as graph neural networks or graph transformers, enhancing their performance on various tasks. For large-scale graphs, where computing PSE for the entire graph may be computationally prohibitive, we thereby sample the neighborhood structure around each node and compute the PSE for these localized subgraphs, which is efficiently parallelized, enabling scalable and efficient generation of PSE for large graphs.

\xhdr{Application on Text-attributed Graphs} Text-attributed graphs consist of nodes associated with textual descriptions and edges representing their relations. Language models are commonly employed to capture both textual and structural information. In this setting, GFSE can be seamlessly integrated to inject structural awareness into textual representations. Given the generated $\mathbf{P}^L \in \mathbb{R}^{N \times d_e}$, a lightweight MLP is trained to project $\mathbf{P}^L$ into the embedding space of the language model. The projected PSEs are prepended as soft tokens to the textual inputs, thereby enriching them with graph topology cues. These structure-augmented tokens are then processed by downstream LLMs, improving their understanding and reasoning on graph-related tasks. The approach, involving training a lightweight MLP and fine-tuning LLM with Parameter-Efficient Fine-Tuning (PEFT) techniques such as LoRA~\cite{hu2021lora}, makes it scalable and efficient for large-scale text-attributed graph applications.





\subsection{Computational Complexity}\label{sec:complexity}
GFSE's complexity comprises two parts: pre-computation of self-supervision labels and pre-training. Computing the shortest path distance between node pairs results in the time complexity of $\mathcal{O}(|E|+|V|\log|V|)$ with node set $V$ and edge set $E$. A brute-force implementation of the subgraph isomorphism counting of fixed size $t$ is $\mathcal{O}(|V|^t)$. We consider the graphlets with at most 5 nodes. One can also choose special graphlet types, \eg paths, cycles, and triangles, which can be efficiently enumerated~\cite{giscard2019general}. Approximating and scalable algorithms can be further used to accelerate this pre-processing step~\cite{fu2024desco,ying2020neural,pashanasangi2020efficiently}. For pre-training, the complexity is $\mathcal{O}(|V|^2)$ for full attention computation and $\mathcal{O}(d|V||E|)$ for initial encoding computation. See runtime evaluations in Appendix~\ref{appd:efficiency}.


\section{Experiments}
We first evaluate the pre-training performance in Sec.~\ref{sec:pretraining} and empirically assess the expressiveness of GFSE on synthetic datasets in Sec.~\ref{sec:expressiveness}. We then evaluate GFSE in a wide range of downstream graph learning tasks in Sec.~\ref{sec:downstream}. Specifically, we conduct experiments with LLMs on text-attributed graphs in Sec.~\ref{sec:exp_llms}.
\subsection{Pre-Training Setup}\label{sec:setting}
\xhdr{Dataset} We utilize diverse datasets for pretraining, ensuring a broad spectrum of graph structures and scales, including MolPCBA, MolHIV, MNIST, peptides, ogbn-proteins, Pokec, ogbn-arxiv and ogbn-product~\cite{wu2018moleculenet,Bhatia16, mikolov2013distributed,szklarczyk2019string, chiang2019cluster, takac2012data,dwivedi2023benchmarking,dwivedi2022LRGB}. These datasets cover several real-world graph domains, such as social networks, academic networks, \textit{etc}. We refer to Table~\ref{table:dataset_info} in Appendix~\ref{app:dataset} for detailed dataset statistics. For large-scale graphs, we first partition them into sets of subgraphs by the METIS algorithm~\cite{karypis1997metis} to handle scalability issues. Training samples from different datasets are mixed and randomly shuffled to form a large-scale pre-training dataset.

\xhdr{Pre-training Setting} The pre-training stage is conducted on the standard train/validation/test splits of the pre-training datasets. The initial encoding dimension $d$ is set as 8. We adopt GIN~\cite{xu2018powerful} as the message-passing layer in the GPS and adopt 8 GPS layers with 8 heads and 128 hidden dimensions. The output dimension is 64 by default. We use Adam as the optimizer with an initial learning rate of $0.001$ and the batch size is 256. The maximum training epochs is 100. An early stopping strategy is used to mitigate overfitting. The pre-training is implemented on the NVIDIA A40 48GB GPU. We refer to Appendix~\ref{app:pre_setting} for more details.

\subsection{Pre-training Evaluation}\label{sec:pretraining}
\begin{figure}[htbp]
\vspace{-0.2cm}
\centering
\includegraphics[width=0.5\textwidth]{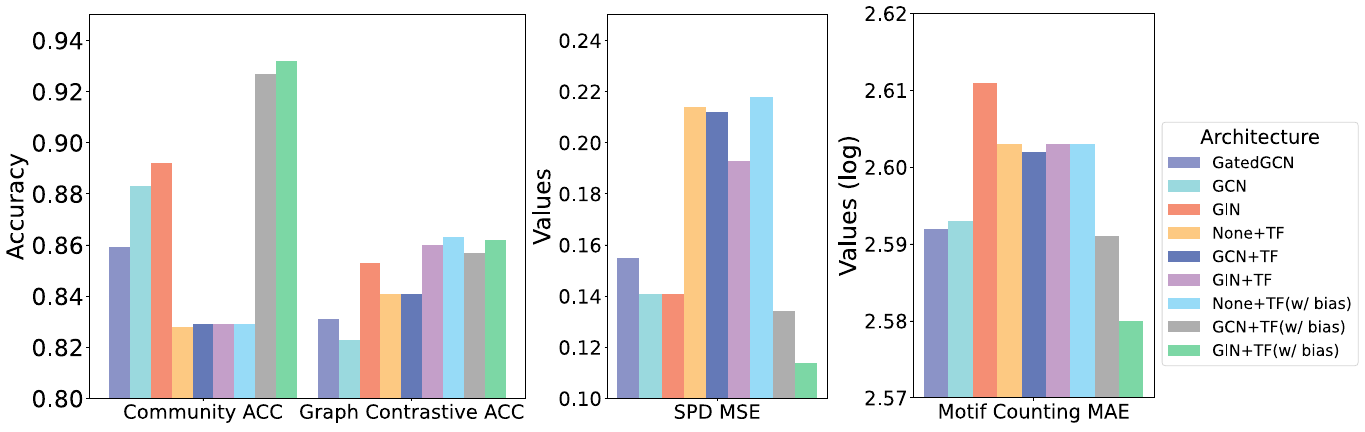}\vspace{-0.3cm}
\caption{Pre-training performance with different architectures. TF stands for transformer.}
\label{fig:archi_task}
\end{figure}

GFSE is pre-trained with four self-supervised learning tasks. We iteratively change the message passing layers (\eg GatedGCN~\cite{bresson2017residual}, GCN~\cite{kipf2016semi} and GIN~\cite{xu2018powerful}) and replace the biased attention with traditional self-attention in the default GFSE architecture. 
We evaluate the pre-training performance with different architectures as shown in Figure~\ref{fig:archi_task}. Accuracy is used to measure community detection and graph contrastive learning tasks, indicating the proportion of node (graph) pairs that are correctly predicted. MSE and MAE are used for shortest path distance and motif counting tasks. See more details in Appendix~\ref{app:metric}. We observe a consistent performance boost when combining biased attention with GIN, which serves as the default configuration of GFSE. 

\subsection{Expressiveness Evaluation }\label{sec:expressiveness}
We empirically evaluate the structure-awareness of the positional and structural encoding (PSE) generated by GFSE on Triangle~\cite{knyazev2019understanding}, Pattern, and Cluster~\cite{dwivedi2023benchmarking} , three datasets that require discerning intricate graph topologies. We evaluate the performance boost brought by GFSE in comparison to traditional RWSE and LapPE. We test on various downstream graph learning models, including MLP, GIN~\cite{xu2018powerful}, transformer~\cite{vaswani2017attention} and GPS~\cite{rampavsek2022recipe}. We further compare with the learning-based PSE approach: GPSE~\cite{liu2023graph}. As shown in Table~\ref{tab:expressivity}, GFSE generates expressive and robust PSE that consistently improves the base model's performance, whereas other structural encodings exhibit considerable variation across different datasets or different base models. Notably, the performance boost brought by GFSE is particularly significant for the Transformer on the Triangle dataset. This demonstrates that GFSE has a stronger enhancement effect on models that originally lack structural bias.
\begin{table}
    \centering
    \caption{Test accuracy ($\%$) enhanced by different positional and structural encoding. The results are averaged over five random seeds. The best results in each dataset are bolded.
    }\vspace{-0.3cm}
    \resizebox{0.9\linewidth}{!}{
    \begin{tabular}{lrrrr}
    \toprule
     & \multicolumn{1}{c}{\textbf{Triangle-S}} & \multicolumn{1}{c}{\textbf{Triangle-L}} & \multicolumn{1}{c}{\textbf{Pattern}} & \multicolumn{1}{c}{\textbf{Cluster}} \\
     \midrule
    MLP+RWSE & 98.22 & 11.88 & 50.53 & 20.96 \\
    MLP+LapPE & 98.60 & 12.62 & 50.53 & 20.96 \\
    MLP+GPSE & 52.80 & 17.42 & 55.66 & 20.96 \\
    \rowcolor{Gray}MLP+GFSE & \textbf{98.71} & \textbf{25.54} & \textbf{57.79} & \textbf{21.28} \\
    \midrule
    GIN & 99.68 & 42.58 & \textbf{85.58} & 60.84 \\
    GIN+RWSE & 99.70 & 40.78 & 85.34 & 61.30 \\
    GIN+LapPE & \textbf{99.74} & 42.48 & 85.45 & 61.83 \\
    GIN+GPSE & 99.32 & 25.32 & 85.19 & 61.95 \\
    \rowcolor{Gray}GIN+GFSE & 99.72 & \textbf{43.84} & \textbf{85.58} & \textbf{63.49} \\
    \midrule
    \multicolumn{5}{l}{Transformer (TF)\textit{ for Triangle and }GPS \textit{for Others}}\\
    TF / GPS & 21.68 & 23.58 & 86.63 & 77.76 \\
    TF / GPS+RWSE & 35.96 & 11.38 & 86.68 & 77.72 \\
    TF / GPS+LapPE & 35.96 & 12.44 & 86.54 & 77.76 \\
    TF / GPS+GPSE & 62.04 & 24.64 & 85.58 & 77.80 \\
    \rowcolor{Gray}TF / GPS+GFSE & \textbf{92.82} & \textbf{30.15} & \textbf{87.98} & \textbf{77.86} \\
    \bottomrule
    \end{tabular}}
    \label{tab:expressivity}
    \vspace{-0.3cm}
\end{table}

\subsection{Downstream Evaluation}\label{sec:downstream}
\subsubsection{Evaluation on Pre-training Domains}
We conduct a comprehensive evaluation of GFSE on real-world graph datasets. We adhere to the experimental setting and hyper-parameters established by previous works~\cite{rampavsek2022recipe} to implement base models. We augment the initial node features with GFSE and evaluate the performance in downstream tasks, comparing it against established structural encoding methods including  RWSE, LapPE, and GPSE~\cite{liu2023graph}. Evaluation results on citation networks (\eg ogbn-Arxiv, PubMed), and image graphs (\eg MNIST and CIFAR10) are given in Table~\ref{exp2}, where PubMed and CIFAR10 are out of the pre-training dataset. More results are provided in Table~\ref{exp1} in Appendix~\ref{appd:downstream_evaluation_performance}.

\begin{table}[htbp]
    \centering
    \caption{Test Accuracy $(\%)$ on ogbn-Arxiv, PubMed, MNIST and CIFAR10.}\label{exp2}\vspace{-0.3cm}
    \resizebox{0.45\textwidth}{!}{
    \begin{tabular}{lcccc} \toprule
 & \textbf{ogbn-arxiv} & \textbf{PubMed} & \textbf{MNIST} & \textbf{CIFAR10} \\
 \midrule
GateGCN & $71.69_{\pm 0.21}$ & $76.86_{\pm 0.41}$ & $97.34_{\pm 0.14}$ & $67.31_{\pm 0.31}$ \\
GateGCN+LapPE & $71.95_{\pm 0.37}$ & $74.83_{\pm 0.24}$ & $97.10_{\pm 0.28}$ & $65.08_{\pm 0.26}$ \\
GateGCN+RWSE & $71.83_{\pm 0.65}$ & $76.11_{\pm 0.39}$ & $96.84_{\pm 0.27}$ & $65.26_{\pm 0.68}$ \\
GateGCN+GPSE & $72.17_{\pm 0.42}$ & $71.97_{\pm 0.36}$ & $96.94_{\pm 0.17}$ & $65.63_{\pm 0.27}$ \\
\rowcolor{Gray} \textbf{GateGCN+GFSE} & $\mathbf{72.61}_{\pm 0.53}$& $\mathbf{78.39}_{\pm 0.84}$& $\mathbf{97.44}_{\pm0.31}$ & $\mathbf{68.39}_{\pm0.47}$ \\
\midrule
Transformer (TF) & $5.86_{\pm 0.00}$ & $\mathbf{66.63}_{\pm 0.73}$ & $97.29_{\pm 0.11}$ & $69.04_{\pm 0.28}$ \\
TF+LapPE & $5.86_{\pm 0.00}$ & $66.27_{\pm 0.46}$ & $96.95_{\pm 0.38}$ & $69.01_{\pm 0.61}$ \\
TF+RWSE & $5.86_{\pm 0.00}$ & $64.43_{\pm 0.37}$ & $97.81_{\pm 0.58}$ & $70.70_{\pm 0.45}$ \\
TF+GPSE & $21.56_{\pm 2.74}$ & $65.89_{\pm 0.14}$ & $97.78_{\pm 0.32}$ & $69.57_{\pm 0.16}$ \\
\rowcolor{Gray} \textbf{TF+GFSE} & $\mathbf{23.84}_{\pm 3.15}$ & $66.30_{\pm 0.68}$ & $\mathbf{98.03}_{\pm 0.84}$ & $\mathbf{71.33}_{\pm 0.23}$ \\ \midrule
GPS & $70.68_{\pm 0.71}$ & $74.26_{\pm 0.60}$ & $98.05_{\pm 0.12}$ & $71.49_{\pm 0.35}$ \\
GPS+LapPE & $69.51_{\pm 0.38}$  & $73.68_{\pm 0.37}$ & $98.16_{\pm 0.28}$ & $71.87_{\pm 0.21}$ \\
GPS+RWSE & $72.14_{\pm 0.84}$ & $72.87_{\pm 0.44}$ & $\mathbf{98.19}_{\pm 0.30}$ & $71.30_{\pm 0.33}$ \\
GPS+GPSE & $71.21_{\pm 0.34}$ & $73.71_{\pm 0.70}$ & $98.08_{\pm 0.13}$ & $72.31_{\pm 0.25}$ \\
\rowcolor{Gray} \textbf{GPS+GFSE} & $\mathbf{72.30}_{\pm 0.13}$ & $\mathbf{75.13}_{\pm 0.35}$ & $98.15_{\pm 0.46}$ & $\mathbf{74.11}_{\pm 0.93}$ \\
\bottomrule
\end{tabular}}\vspace{-0.3cm}
\end{table}

\xhdr{Results} From Table~\ref{exp2}, we observe that the integration of GFSE achieves state-of-the-art performance in 10 out of 12 evaluated cases. This enhancement is not confined to a specific domain; it is evident across diverse tasks, including node classification on large-scale citation networks and graph-based image classification. Notably, GFSE consistently outperforms both traditional hand-crafted encodings, which show inconsistent benefits, and the existing pre-trained baseline, GPSE. The dramatic performance leap observed when applying GFSE to the standard Transformer is particularly telling, as it highlights our model's ability to inject essential structural inductive biases into architecturally general models, enabling them to effectively reason over complex graph data.
\subsubsection{Evaluation on Molecular Graphs} \label{sec:molecule}
We use the molecular datasets, namely Tox21, Sider, BBBP, and MUV~\cite{hu2020open} as a downstream task. We evaluate the effectiveness of GFSE under two settings:
(1) In the training-from-scratch setting, we directly concatenate GFSE's PSE with the raw node features to create new input features. This augmented representation is then fed into a randomly initialized model from the beginning of training. We take GINE~\cite{xu2018powerful} and GPS~\cite{rampavsek2022recipe} as our backbone. (2) In the fine-tuning setting, we assess GFSE's ability to enhance pre-trained models by concatenating the node encodings obtained from a pre-trained model with the GFSE's PSE. The concatenated features are then fed into the final read-out layers for prediction. During fine-tuning, the parameters of the pre-trained model and read-out layers are updated. We select GraphMAE~\cite{hou2022graphmae} and MoleBERT~\cite{xia2022mole} as the pre-trained backbones and compare with other baselines without structural encoding, namely SSP~\cite{hu2019strategies}, GraphLoG~\cite{xu2021self}, GraphCL~\cite{you2020graph}.
Refer to Appendix~\ref{app:evaluation_molecule} for more implementation details.

\xhdr{Results} Experimental results are shown in Table.~\ref{tab:molecule_data}. For training the models from scratch, on both GINE and GPS, PSE consistently improves model performance, achieving better results than all the other structural feature augmentation methods. As to fine-tuning,  GFSE significantly boosts the performance and achieves state-of-the-art performance.   

\begin{table}[h]\vspace{-0.4cm}
\centering
\caption{Test ROC-AUC(\%) performance on molecular datasets. The best results for each dataset are bolded.}
\label{tab:molecule_data}\vspace{-0.3cm}
\resizebox{0.9\linewidth}{!}{
    \begin{tabular}{lcccc}
    \toprule
     & \textbf{Tox21} & \textbf{Sider} & \textbf{BBBP} & \textbf{MUV} \\ \midrule
     SSP & 76.8 $\pm$ 0.8 & 61.7 $\pm$ 0.8 & 67.9 $\pm$ 0.9 & 79.8 $\pm$ 1.6 \\
     GraphCL & 75.7 $\pm$ 0.5 & 60.8 $\pm$ 0.7 & 69.5 $\pm$ 0.5 & 74.5 $\pm$ 1.3 \\
     GraphLoG & 75.4 $\pm$ 0.9 & 61.2 $\pm$ 1.1 & 72.5 $\pm$ 0.8 & 76.0 $\pm$ 1.1 \\ \midrule
    \multicolumn{5}{l}{\textit{\textbf{Train-from-scratch Models}}}\\
     GINE & 74.5 $\pm$ 0.4 & 58.6 $\pm$ 0.1 & 67.7 $\pm$ 0.7 & 74.8 $\pm$ 0.6 \\
     GINE+RWSE & 75.3 $\pm$ 0.2 & 58.4 $\pm$ 1.8 & 66.7 $\pm$ 1.4 & 76.4 $\pm$ 0.8 \\
     GINE+LapPE & \textbf{77.6 $\pm$ 0.8} & 57.2 $\pm$ 1.1 & 65.8 $\pm$ 0.3 & 77.0 $\pm$ 0.8 \\
     GINE+GPSE & 74.9 $\pm$ 0.4 & 60.1 $\pm$ 0.8 & 66.4 $\pm$ 0.1 & 75.8 $\pm$ 1.3 \\
     \cellcolor{Gray}GINE+GFSE & \cellcolor{Gray}75.5 $\pm$ 0.7 & \cellcolor{Gray}\textbf{60.9 $\pm$ 0.5} & \cellcolor{Gray}\textbf{69.1 $\pm$ 1.3} & \cellcolor{Gray}\textbf{77.7 $\pm$ 1.2} \\ \hdashline
     GPS & 73.9 $\pm$ 0.2 & 58.6 $\pm$ 0.4 & 67.1 $\pm$ 0.3 & 68.0 $\pm$ 0.6 \\
     GPS+RWSE & 74.6 $\pm$ 1.3 & 56.4 $\pm$ 0.6 & 67.9 $\pm$ 1.0 & 69.7 $\pm$ 0.6 \\
     GPS+LapPE & 74.8 $\pm$ 1.1 & 60.5 $\pm$ 0.6 & 67.9 $\pm$ 0.6 & 70.1 $\pm$ 2.2 \\
     GPS+GPSE& 75.1 $\pm$ 0.7 & 56.6 $\pm$ 1.7 & 67.8 $\pm$ 0.7 & 68.3 $\pm$ 0.1 \\
     \cellcolor{Gray}GPS+GFSE & \cellcolor{Gray}\textbf{76.3 $\pm$ 1.4} & \cellcolor{Gray}\textbf{61.8 $\pm$ 0.5} & \cellcolor{Gray}\textbf{68.0 $\pm$ 0.5} & \cellcolor{Gray}\textbf{73.6 $\pm$ 0.5} \\ 
    \midrule 
     \multicolumn{5}{l}{\textit{\textbf{Pre-trained Models}}} \\
     GraphMAE & 75.4 $\pm$ 0.4 & 59.8 $\pm$ 0.5 & 69.5 $\pm$ 1.6 & 76.3 $\pm$ 2.4 \\
     GraphMAE+RWSE & \textbf{76.3 $\pm$ 0.5} & 60.5 $\pm$ 0.8 & 66.4 $\pm$ 3.7 & 77.7 $\pm$ 1.5 \\ 
     \cellcolor{Gray}GraphMAE+GFSE & \cellcolor{Gray}75.9 $\pm$ 0.9 & \cellcolor{Gray}\textbf{62.1 $\pm$ 0.8} & \cellcolor{Gray}\textbf{70.5 $\pm$ 1.4} & \cellcolor{Gray}\textbf{78.1 $\pm$ 1.3} \\ \hdashline
     MoleBERT & 76.8 $\pm$ 0.5 & 62.8 $\pm$ 1.1 & \textbf{71.9 $\pm$ 1.6} & 78.6 $\pm$ 1.8 \\
     MoleBERT+RWSE & 77.8 $\pm$ 0.7 & \textbf{63.1 $\pm$ 0.6} & 66.5 $\pm$ 2.1 & 80.4 $\pm$ 1.3 \\
     \cellcolor{Gray}MoleBERT+GFSE & \cellcolor{Gray}\textbf{78.0 $\pm$ 0.4} & \cellcolor{Gray}\textbf{63.1 $\pm$ 0.7} & \cellcolor{Gray}68.9 $\pm$ 2.1 & \cellcolor{Gray}\textbf{80.5 $\pm$ 2.0} \\  
    \bottomrule
    \end{tabular}
}
\end{table}\vspace{-0.4cm}
\begin{table}[h]
\centering
\caption{Evaluation on text-attributed graphs}
\label{tab:LLM}\vspace{-0.3cm}
\resizebox{\linewidth}{!}{
    \begin{tabular}{lcc|cc|cc|c|c}
    \toprule
     & \multicolumn{2}{c|}{\textbf{Cloth}} & \multicolumn{2}{c|}{\textbf{Home}} & \multicolumn{2}{c|}{\textbf{Sport}} & \textbf{Arxiv} & \textbf{Cora} \\
    \textbf{} & Hit@1 & MRR & Hit@1 & MRR & Hit@1& MRR & ACC& ACC \\
    \midrule
    InstructGLM & 76.23 & 82.60 & 79.82 & 85.93 & 62.50 & 73.25 &75.70 & 87.08\\  
    \midrule
    Finetuned LLaMA & 74.73 & 82.87 & 78.93 & 86.07 & 62.52 & 75.77 &74.94 &79.95 \\  
    \midrule
    \quad + RRWP & 76.08& 84.11 & 74.64 & 85.24& 62.71 &75.20& 76.03 & 90.27\\  \midrule
    \quad + GraphSAGE & 76.22 & 84.16 & 73.74 & 81.66 & 62.26 & 75.36 &  \textbf{76.13} & 90.83\\  
    \midrule
    \quad + \textbf{GFSE (Ours)} & \textbf{76.84} & \textbf{84.68} & \textbf{79.85} & \textbf{86.77} & \textbf{64.79} & \textbf{76.24} & 76.10 & \textbf{91.26}\\  
    \bottomrule
    \end{tabular}
    }\vspace{-0.3cm}
\end{table}
\subsection{Integration with Large Language Models}\label{sec:exp_llms}
We evaluate GFSE on the task of link prediction using three text-attributed graphs: the citation networks Arxiv~\cite{hu2020open} and Cora~\cite{yang2016revisiting}, and the e-commerce network Amazon-Product~\cite{he2016ups, mcauley2015image}. In these datasets, each node (\eg a paper or product) features a detailed textual description, and edges signify relations such as co-authorship or co-purchase. To synergize our structural encodings with an LLM, we employ a lightweight MLP to project the 32-dimensional PSE into the 4096-dimensional embedding space of LLaMA2-7B~\cite{touvron2023llama}, aligning the structural and textual modalities. The input for the LLM is constructed by concatenating the central node’s text with that of its one-hop neighbors. This sequence is then prepended with the projected PSE as a soft structural token and appended with a special graph token. The final hidden state corresponding to this special token serves as the comprehensive node representation. Following a prior setup~\cite{zhu2024parameter}, we train the model to predict edge likelihoods based on the cosine similarity between node-pair representations. Specifically, we train the MLP projector and fine-tune the LLM using LoRA~\cite{hu2021lora} with the contrastive objective: $\mathcal{L}=\sum_{(i,j)\in E \& (i,j')\notin E}\big(d_{ij}^2+\max(\tau-d_{ij'},0)^2\big)$,
where $d_{ij}=1-\cos(v_i,v_j)$, $v_i$ denotes the $i$-th node representation, and $\tau=0.5$ is the margin.

\xhdr{Results} Evaluation results on text-attributed graphs are reported in Table~\ref{tab:LLM}. We select InstructGLM~\cite{ye2023natural} as a baseline, which has been fine-tuned on graph domains without structural information. Additionally, we include comparisons against RRWP~\cite{ma2023graph}, and GraphSAGE~\cite{hamilton2017inductive}, which are trained from scratch as a PSE encoder with LLM finetuning together. \textit{Finetuned LLaMA} refers to the LLaMA model~\cite{touvron2023llama} fine-tuned without any PSE. As seen from the table, GFSE, which is pre-trained on cross-domain graph data as a structural encoder, outperforms other methods in general. In particular, GFSE achieves an average performance boost of $2.66\%$ over InstructGLM, $2.02\%$ over GraphSAGE encoder and $1.53$ over RRWP across the five datasets. These gains demonstrate the effectiveness of GFSE in language-based graph tasks. See more discussions in Appendix~\ref{appd:downstream_evaluation_performance}.

\subsection{More Experimental Analysis} 
\xhdr{Reconstruction of other PSE types} The pre-trained GFSE is able to generate effective PSE, which can reconstruct other pre-defined PSE, such as LapPE~\cite{lim2022sign} and ElstaticPE~\cite{kreuzer2021rethinking}. Table~\ref{tab:reconstruction} demonstrates that the PSE generated by the pre-trained GFSE, followed by a trainable lightweight MLP, is capable of reconstructing various other pre-defined PSEs. We evaluate this using the coefficient of determination $R^2$ scores as a metric. Notably, our model performs competitively compared with GPSE, given the fact that GPSE directly adopts PSE reconstruction as its training objective. Instead, our method generalizes well across different PSEs without being directly trained for reconstruction. This suggests that the structural self-supervision pre-training tasks are effective and sufficient in capturing important structural information.
\begin{table}[htbp]\vspace{-0.3cm} 
\centering 
\caption{Performance of other PSE reconstruction. The coefficient of determination $R^2$ scores are reported as the metric.} \label{tab:reconstruction}\vspace{-0.4cm}
\resizebox{0.45\textwidth}{!}{
\begin{tabular}{llllll} \toprule
PSE type & ElstaticPE & LapPE & RWSE & HKdiagSE & CycleSE \\ \midrule
GPSE & \textbf{0.964} & 0.973 & 0.984 & 0.981 & 0.977 \\
Ours & 0.947 & \textbf{0.986} & \textbf{0.987} & \textbf{0.984} & \textbf{0.992} \\
\bottomrule
\end{tabular}}\vspace{-0.4cm}
\end{table}

\xhdr{Ablation on Tasks and Model Architecture} We analyze the sensitivity of each pre-training task and model architecture \wrt the performance boost in downstream tasks. The results are illustrated in Table~\ref{tab:ablation}. Firstly, we iteratively remove one pre-training task and follow the same setting to pre-train and evaluate GFSE. We observe that each removal results in a discernible reduction in downstream performance. Notably, the shortest path distance task is particularly critical for the ZINC and CIFAR10 datasets, while community detection is significant for academic datasets. We also conduct ablation studies on the main components of GFSE. Specifically, we use traditional attention to replace biased attention and remove GIN or attention modules, respectively. 
We notice that all the above architecture variants lead to performance degradation. The hybrid approach outperforms both attention-only and GIN-only setups, suggesting that integrating biased attention mechanisms can compensate for the absence of global information in local message-passing layers.

\begin{table}[h]
\vspace{-0.4cm}  
\captionof{table}{Ablation studies on the pre-training tasks and model architecture of GFSE. Best results are shown in bold.}
\label{tab:ablation} \vspace{-0.3cm}
\resizebox{0.9\linewidth}{!}{
    \begin{tabular}{lccc}\toprule
     &\textbf{ZINC} & \textbf{CIFAR10}& \textbf{Arxiv} \\ 
     & \textbf{MAE} $\downarrow$& \textbf{ACC} $\uparrow$& \textbf{ACC} $\uparrow$\\ \midrule
    GPS w/o PE& 0.1182 & 71.49 & 70.68 \\
    Augment by GFSE & \textbf{0.0613} & \textbf{74.11} & 72.30 \\ \hdashline
    \multicolumn{4}{l}{\textbf{\textit{Different Pre-training Tasks}}} \\
    \quad \textit{w/o} Community Detection & 0.0637 & 72.38 & 70.34 \\
    \quad\textit{w/o} Motif Counting & 0.0731 & 73.02 & 71.27 \\
    \quad\textit{w/o} Shortest Path Distance Regression & 0.1074 & 71.58 & 72.06 \\
    \quad\textit{w/o} Graph Contrastive Learning & 0.0856 & 73.02 & 72.13 \\ \hdashline
    \multicolumn{4}{l}{\textbf{\textit{Different Model Architecture}}} \\
    \quad GIN+Traditional Attention & 0.0872 & 73.13& 71.85 \\
    \quad Biased Attention Only & 0.1137 & 70.97& 71.33 \\
    \quad GIN Only  & 0.0640 &72.31 & \textbf{72.34} \\
    \bottomrule
    \end{tabular}} \vspace{-0.2cm}
\end{table}

\begin{figure}[htbp]\vspace{-0.3cm} 
  \centering
  \includegraphics[width=0.3\textwidth]{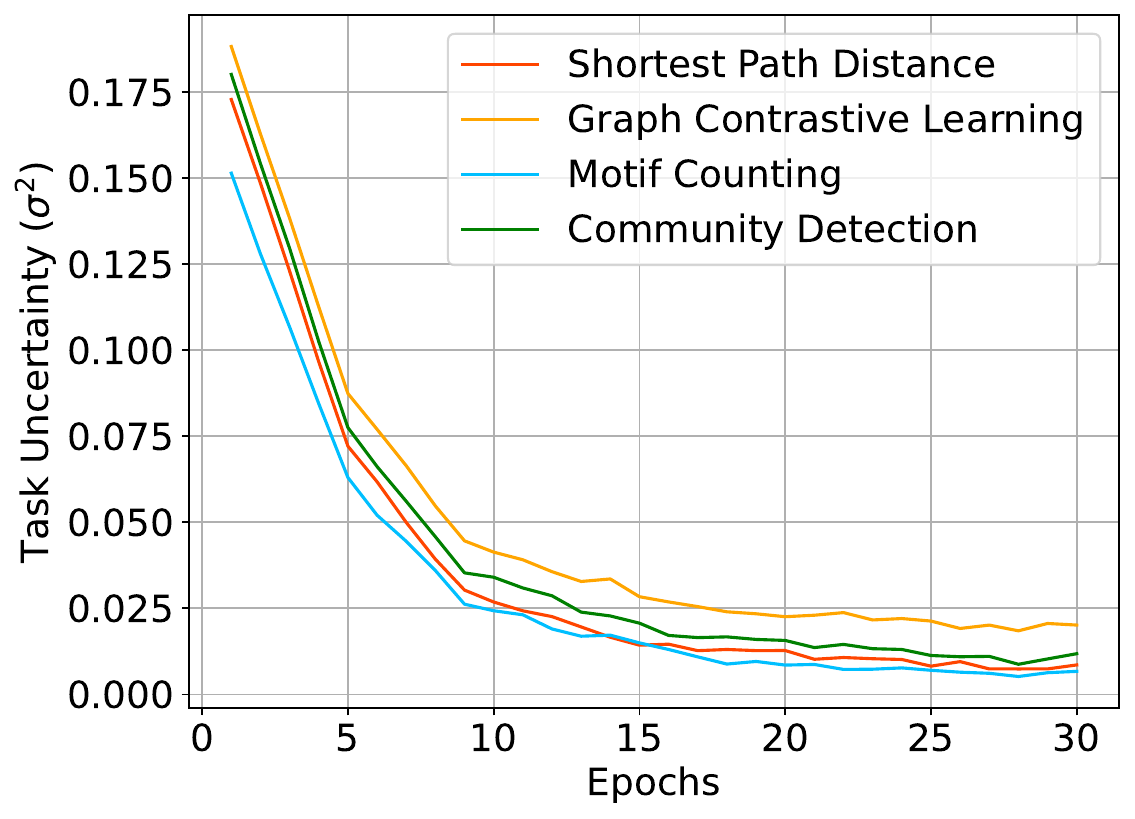} \vspace{-0.4cm} 
   \caption{Learning task uncertainty ($\sigma^2$) \wrt pre-training}
  \label{fig:task_uncertainty} \vspace{-0.2cm} 
\end{figure}

\xhdr{Task-specific Uncertainty} Figure~\ref{fig:task_uncertainty} illustrates the trajectory of task uncertainty ($\sigma^2$) across different pre-training tasks \wrt pre-training epochs. Higher values of $\sigma^2$ reduce the respective task's contribution to the overall training loss. We observe that all tasks show a sharp decline in uncertainty during the pre-training process. Notably, motif counting maintains a lower uncertainty throughout the training process compared to other tasks, suggesting that it might be inherently more straightforward for the model to optimize or more integral to the model's overall learning structure.

\begin{figure}[htbp]
  \centering
  \vspace{-0.3cm}  
  \includegraphics[width=0.4\textwidth]{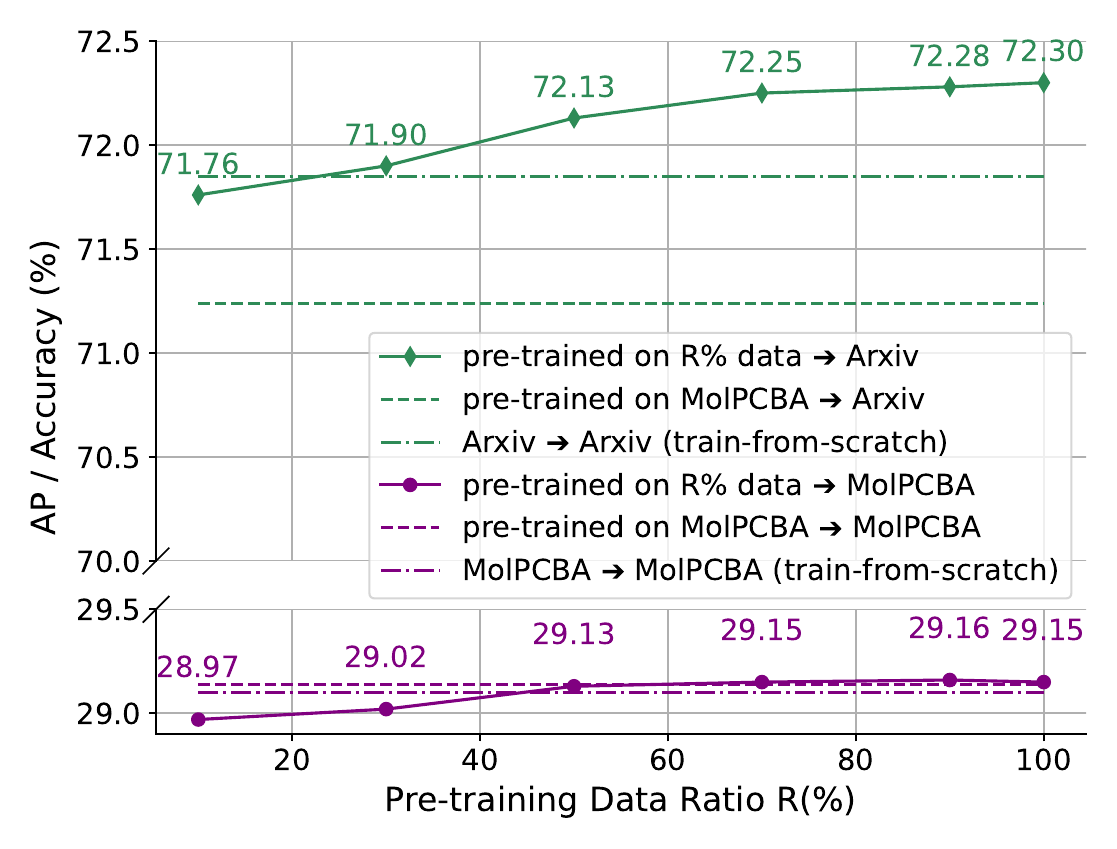} \vspace{-0.3cm}
   \caption{Scalability of GFSE with GPS as a backbone.}
  \label{fig:pre-training}
  \vspace{-0.5cm}
\end{figure}

\xhdr{Impact of Pre-training Data Size} Figure~\ref{fig:pre-training} illustrates the impact of pre-training data scale and domain transfer on GFSE’s performance, revealing several key insights. First, as the pre-training data ratio increases, the model’s performance consistently improves, indicating that larger and more diverse pre-training datasets are essential for capturing richer structural representations and achieving stronger generalization. Second, models initialized with GFSE pre-training outperform their train-from-scratch counterparts, demonstrating the effectiveness of structural pre-training in providing transferable inductive biases that enhance downstream adaptation. Lastly, cross-domain pre-training proves to be important for domain generalization. GFSE pre-trained on a single dataset struggles to generalize to different domains (\eg from MolPCBA to Arxiv). This indicates that models trained on a single domain may overfit to the specific characteristics, limiting their generalization to other tasks. In contrast, cross-domain pre-training enables the model to learn more robust, domain-invariant representations, crucial for scaling to diverse real-world text-attributed graph applications.

\section{Conclusion}

GFSE represents a meaningful step toward a universal graph encoder, leveraging multiple self-supervised objectives and relative positional encoding within a Graph Transformer to capture expressive and transferable structural patterns across diverse graph domains. Extensive experiments on both synthetic and real-world datasets demonstrate that GFSE consistently enhances the performance of downstream graph encoders across a wide range of tasks. The effectiveness of GFSE can be influenced by the quality and diversity of the pre-training graphs, as biases or under-representation in these data sources may affect the learned representations. Incorporating richer, more heterogeneous, and potentially dynamic graph data is a natural direction for future research. More broadly, GFSE holds promise for real-world applications such as computational biology, social network analysis, and recommendation systems, while reducing reliance on heavy task-specific fine-tuning and improving the accessibility of strong graph representations under limited computational budgets.

\newpage
\bibliographystyle{ACM-Reference-Format}
\bibliography{reference}

\appendix
\appendix

\section{Methodologies}\label{app:experiment}
\subsection{Dataset}\label{app:dataset}
\begin{table}[htbp]\vspace{-0.4cm}
\centering
\caption{Dataset Information. \textit{class.} represents classification task and \textit{reg.} represents regression task.}
\label{table:dataset_info}
\resizebox{0.5\textwidth}{!}{
\begin{tabular}{@{}lrrrccrc@{}}
\toprule
\multirow{2}{*}{Dataset} & \multicolumn{1}{c}{Num.} & \multicolumn{1}{c}{Num.} & \multicolumn{1}{c}{Num.} & \multicolumn{1}{c}{Pred.} & \multicolumn{1}{c}{Pred.} & \multicolumn{1}{c}{Num.} & \multirow{2}{*}{Metric} \\
                         & \multicolumn{1}{c}{graphs} & \multicolumn{1}{c}{nodes}  & \multicolumn{1}{c}{edges}  & \multicolumn{1}{c}{level} & \multicolumn{1}{c}{task} & \multicolumn{1}{c}{tasks} & \\ \midrule
MolPCBA                  & 437,929  & 25.97  & 28.11  & graph & class. (binary) & 128 & AP \\
MolHIV                   & 41,127   & 25.51  & 27.46  & graph & class. (binary) & 1   & AUROC \\
MNIST                    & 70,000   & 70.57  & 281.65 & graph & class. (10-way) & 1   & ACC \\
Peptides-func            & 15,535   & 150.94 & 153.65 & graph & class. (binary) & 10  & AP \\
Peptides-struct          & 15,535   & 150.94 & 153.65 & graph & reg.            & 11  & MAE \\
ogbn-proteins            & 1        & 132,534  & 39,561,252 & node & class. (binary) & 112 & AUROC \\
Pokec                    & 1 & 1,632,803 & 30,622,564 & node & class. (binary) & 1 & ACC \\
ogbn-arxiv               & 1 & 169,343 & 1,166,243 & node & class. (40-way) & 1 & ACC \\
ogbn-products            & 1 & 2,449,029 & 61,859,140 & node & class. (47-way) & 1 & ACC \\
ZINC & 249,456 & 23.2 & 49.8 & graph & reg. & 1 & MAE \\
PubMed & 19,717 & 88,648 & 500 & node & class. (3-way) & 1 & ACC \\
CIFAR10 & 60,000 & 117.6 & 941.2 & graph & class. (10-way) & 1 & ACC \\
\bottomrule
\end{tabular}}\vspace{-0.4cm}
\end{table}

\begin{table}[htbp]
    \centering
    \caption{Dataset statistics of three categories from Amazon e-commerce networks}\label{tab:tag_stat}
\resizebox{0.5\textwidth}{!}{
\begin{tabular}{lcccc} \toprule
 & \textbf{\# nodes} & \textbf{\# edges} & \textbf{avg. degree} & \textbf{avg. \# tokens} \\ \midrule
\textbf{Clothing} & 469,274 & 2,578,746 & 10.99 & 117.83 \\
\textbf{Home} & 453,121 & 3,732,948 & 16.48 & 133.94 \\
\textbf{Sports} & 293,712 & 2,390,076 & 16.27 & 125.08 \\
\bottomrule
\end{tabular}}\vspace{-0.4cm}
\end{table}
\subsection{Pre-training Metric}\label{app:metric}
We use an accuracy metric to measure Community Detection and Graph Contrastive Learning and use mean squared error (MSE) to measure the performance of Shortest Path Distance regression and Motif Counting. 
\begin{itemize}[leftmargin=*]
    \item For the community detection task, we set $\epsilon$ as $1$. The predicted label between the $i$-th node and the $j$-th node $\hat{Y}_{i,j}^c$ is $1$ if $\operatorname{sim}(i,j)\geq 0.5$ and $0$ otherwise. Accuracy is calculated by comparing the predicted label $\hat{Y}_{i,j}^c$ with the ground truth labels $Y_{i,j}^c$ and is defined as the proportion of correctly predicted labels out of all possible node pairs:
    \begin{equation}
        \text{Accuracy(CD)} = \frac{\sum_{(i,j)\in V}\mathbbm{1}(\hat{Y}_{i,j}^c=Y_{i,j}^c)}{|V|(|V|-1)/2},
    \end{equation}
    where $\mathbbm{1}(\cdot)$ is an indicator function and $|V|(|V|-1)/2$ is the total number of unique node pairs in the graph. This metric effectively measures how well the model can identify community structures by correctly classifying node pairs as being in the same community or in different communities.
    \item For the graph contrastive learning task, we evaluate pre-training performance using the accuracy metric, which measures the model's ability to correctly classify graphs as originating from the same or different datasets. The accuracy is computed by:
    \begin{equation}
        \text{Accuracy(GCL)} = \frac{\sum_{i,j}\mathbbm{1}(\hat{Y}_{G_i,G_j}=Y_{G_i,G_j})}{N},
    \end{equation}
    where $\hat{Y}_{G_i,G_j}$ is the predicted label indicating whether graph $G_i$ and $G_j$ are from the same dataset and $Y_{G_i,G_j}$ is the ground truth label. $\hat{Y}_{G_i,G_j}$ is $1$ if $\operatorname{sim}(\boldsymbol{z}_{G_i},\boldsymbol{z}_{G_j})\geq 0$ and $0$ otherwise.$N$ is the total number of evaluated graph pairs. 
    \item For shortest path distance regression, the mean squared error (MSE) is used as a metric, which is defined as:
    \begin{equation}
        \text{MSE(SPD)}  =\frac{1}{|E|}\sum_{(i,j)\in E}(h_{\text{SPD}}(\mathbf{P}^L_i\|\mathbf{P}^L_j)-\text{SPD}_{i,j})^2.
    \end{equation}
    The ground truth SPD is normalized by the graph diameter to ensure scale consistency and training stability.
    \item For the motif counting task, the mean absolute error is used as a metric, which is defined as:
    \begin{equation}
        \text{MAE(MC)} = \frac{1}{|V|}\sum_{i\in V}\|h_{\text{MC}}(\mathbf{P}_i^L)-Y_i^m\|_1,
    \end{equation}
    where $Y_i^m$ is the pre-computed label for the $i$-th node.
\end{itemize}

\subsection{Uncertainty-based Loss Weighing}\label{app:uncertainty}
The scale of the loss of different tasks can be different, causing the overall loss to be dominated by a certain task, and ultimately the loss of the other tasks cannot affect the learning process of the network-sharing layers. We use the uncertainty-based loss-weighing method~\cite{kendall2018multi} to automatically balance the four pre-training tasks and unify the different scales. Moreover, the uncertainty value reflects the contribution of each task towards the overall pre-training process. A higher uncertainty value indicates a lower contribution~\cite{kendall2018multi}. Let $\sigma_\tau$ and $\mathcal{L}_\tau$ represent the task-specific uncertainty value for the task $\tau$. The overall pre-training loss is computed by:
\begin{equation}
\begin{aligned}
    \mathcal{L} =& \frac{1}{\sigma_{\text{SPD}}^2}\mathcal{L}_{\text{SPD}} +\frac{1}{\sigma_{\text{MC}}^2}\mathcal{L}_{\text{MC}} + \frac{1}{\sigma_{\text{CD}}^2}\mathcal{L}_{\text{CD}} + \frac{1}{\sigma_{\text{GCL}}^2}\mathcal{L}_{\text{GCL}} \\
    & + \log \sigma_{\text{SPD}} +\log \sigma_{\text{MC}}+\log \sigma_{\text{CD}}+\log \sigma_{\text{GCL}}.
\end{aligned}
\end{equation}

\subsection{Pre-training Setting}\label{app:pre_setting}
The pre-training stage is conducted on the standard train/validation/test splits of the pre-training datasets. The dimension of initial encoding $d$ is set as 8. We try GatedGCN~\cite{bresson2017residual}, GCN~\cite{kipf2016semi} and GIN~\cite{xu2018powerful} as the message-passing layers in the GPS. The number of GPS layers is tuned in the range of [4, 16] and the number of heads is tuned within $\{4,8, 16\}$. The hidden dimension is tuned $\{32, 64, 128, 256\}$. The output PSE dimension is in $\{32, 64\}$. The temperature $\tau$ is set as $0.1$ and the margin$\epsilon$ is 0. We use Adam as the optimizer with an initial learning rate of $0.001$, and the batch size is set as 256. The maximum training epochs is 100. An early stopping strategy is used to mitigate overfitting. The experiments are implemented on the NVIDIA A40 48GB GPU.

\section{More Experimental Results}\label{appd:performance}
\subsection{Integration with Pre-trained Models}\label{app:evaluation_molecule}
All models are fine-tuned, trained or tested using five different seeds from 42 to 46, with the results averaged. Additionally, for the results of our generated PSE, we select three different seeds to obtain three GFSE checkpoints. Each GFSE is used to run the downstream task five times with the aforementioned seeds ($42-46$), and all results are averaged. 

For the training from scratch setting, we adopt and modify the code base from GPS~\cite{rampavsek2022recipe} \footnote{\url{https://github.com/rampasek/GraphGPS}}. RWSE and LapPE are of dimension $32$ for molecule benchmark in Table~\ref{tab:molecule_data} across all the datasets. Given a graph $G(V, E)$, we directly concatenate the PSE and the raw node feature as the new input feature, then send them into the very beginning of a model with randomly initialized parameters, which is as follows:
\begin{equation*}
 X ' = \operatorname{concat}(X, \text{PSE});\quad  \hat{y} =  \operatorname{MLP}[\operatorname{pooling}, [\operatorname{GraphModel} (X')]]
\end{equation*}
where $X$ denotes the raw node feature of the input graph, $X'$ is the input feature augmented with structural information, $\operatorname{GraphModel}$ denotes our backbones GNN or GPS, and the read-out layer consists of pooling and MLPs to obtain the final prediction.


For the fine-tuning setting in the molecule benchmark, we concatenate the node encodings obtained from a pre-trained model with the extra structural features. Then we send the concatenated features into the final read-out layers for the final prediction. Note that during the fine-tuning process, the parameters of the entire model (both the pre-trained model and the read-out layer) are continuously updated.
\begin{equation}
 X' = \operatorname{GNN} (X); \quad  \hat{y} =  \operatorname{MLP}[\operatorname{pooling} [\operatorname{concact} (X', \text{PSE})]],
\end{equation}
where $X'$ denotes the latent node features output from the pre-trained models, $\text{PSE}$ denotes the extra structural feature generated by GFSE, the read-out layer consists of pooling and MLPs, where the hyperparameters follow exactly as ~\cite{xia2022mole}.

\subsection{Downstream Evaluation Performance}\label{appd:downstream_evaluation_performance}
\begin{table}[htbp]\vspace{-0.4cm}
\centering
    \caption{Performance on MolPCBA, ZINC (subset), Peptides-func and Peptides-struct. The best results in each dataset are bolded.}
    \resizebox{0.9\linewidth}{!}{
    \begin{tabular}{lcccc}
    \toprule
     & \textbf{MolPCBA} & \textbf{ZINC} & \textbf{Peptides-func} & \textbf{Peptides-struct} \\
     & \textbf{AP} $\uparrow$ & \textbf{MAE} $\downarrow$ & \textbf{AP} $\uparrow$& \textbf{MAE} $\downarrow$\\ \midrule
     GCN & $0.2424_{\pm 0.0034} $ & $0.3670 _{\pm 0.0110}$ & $0.5930_{\pm  0.0023}$ & $0.3496_{\pm 0.0013}$ \\
    GCN+LapPE & $0.2417_{\pm 0.0047}$ & $0.2052_{\pm 0.0132}$ & $0.6021_{\pm 0.0051}$ & $0.2688_{\pm 0.0027}$ \\
     GCN+RWSE & $0.2438_{\pm 0.0028}$ &$ 0.1741_{\pm 0.0528}$ & $0.5827_{\pm 0.0046}$ & $0.3270_{\pm 0.0019}$ \\
     GCN+GPSE & $0.1958_{\pm 0.0074}$ & $\mathbf{0.1218}_{\pm 0.0613}$ & $0.5959_{\pm 0.0034}$ & $0.2710_{\pm  0.0041}$ \\
     \rowcolor{Gray} GCN+GFSE & $\mathbf{0.2477}_{\pm  0.0021}$ & $0.1237_{\pm 0.0428}$ & $\mathbf{0.6131}_{\pm   0.0074}$ & $\mathbf{0.2513}_{\pm  0.0054}$ \\ \midrule
      GIN & $0.2703_{\pm 0.0023}$ & $0.5260_{\pm 0.0510}$ & $0.5498_{\pm 0.0079}$ & $0.3547_{\pm 0.0045}$ \\
       GIN+LapPE & $0.2701_{\pm 0.0013}$ & $0.2203_{\pm 0.0386}$ & $0.5323_{\pm 0.0083}$ & $0.2650_{\pm 0.0041}$ \\
      GIN+RWSE & $0.2781_{\pm 0.0031}$ & $0.1731_{\pm 0.0614} $& $0.5410_{\pm 0.0068}$ & $0.3282_{\pm 0.0037}$ \\
      GIN+GPSE & $0.2765_{\pm 0.0073}$ & $0.2162_{\pm 0.0429}$ & $0.5389_{\pm 0.0094}$ & $\mathbf{0.2581}_{\pm 0.0046}$ \\
    \rowcolor{Gray} GIN+GFSE & $\mathbf{0.2839}_{\pm 0.0046}$ & $\mathbf{0.1689}_{\pm 0.0524}$ & $\mathbf{0.5532}_{\pm 0.0103}$ & $0.2674_{\pm 0.0039}$ \\ 
     \midrule
     Transformer (TF) & $0.0808_{\pm 0.0117}$ & $0.6943_{\pm 0.0328}$ & $0.4800_{\pm 0.0076}$ & $0.4192_{\pm  0.0028}$ \\
      TF+LapPE & $0.1784_{\pm 0.0329}$ & $0.5101_{\pm 0.0724}$ &$ 0.6307_{\pm 0.0091}$ & $0.2514_{\pm 0.0031}$ \\
      TF+RWSE & $0.2083_{\pm 0.0674}$ & $0.2193_{\pm 0.0640}$ & $0.6326_{\pm 0.0028}$ & $0.3344_{\pm 0.0028}$ \\
      TF+GPSE & $0.2040_{\pm 0.0531}$ & $0.1883_{\pm 0.0263}$ & $0.6534_{\pm 0.0041}$ & $0.2479_{\pm 0.0068}$ \\
      \rowcolor{Gray} TF+GFSE & $\mathbf{0.2376}_{\pm 0.0342}$ & $\mathbf{0.1548}_{\pm 0.0796}$ & $\mathbf{0.6642}_{\pm  0.0025}$ & $\mathbf{0.2436}_{\pm   0.0071}$ \\
     \midrule
     GPS & $0.2869_{\pm 0.0045}$ & $0.1182_{\pm 0.0049}$ & $0.6535_{\pm 0.0041}$ & $0.2500_{\pm 0.0012}$ \\ 
      GPS+LapPE & $\mathbf{0.2939}_{\pm 0.0016}$ & $0.1078_{\pm 0.0084}$ & $0.6494_{\pm  0.0037}$ & $0.2501_{\pm 0.0026}$ \\
      GPS+RWSE & $0.2907_{\pm 0.0028}$ & $0.0700_{\pm 0.0040}$ & $0.6603_{\pm 0.0085}$ & $0.2739_{\pm 0.0063}$ \\
      GPS+GPSE & $0.2911_{\pm  0.0036}$ & $0.0648_{\pm 0.0030}$ & $0.6688_{\pm 0.0151}$ & $\mathbf{0.2464}_{\pm 0.0025}$ \\
      \rowcolor{Gray} GPS+GFSE & $0.2916_{\pm 0.0061}$ & $\mathbf{0.0613}_{\pm 0.0026}$ & $\mathbf{0.6874}_{\pm 0.0120}$ & $0.2474_{\pm  0.0051}$ \\ 
      \midrule
      GFSE Imp.($\%$) &$32.60$&$76.43$&$2.78$&$42.47$ \\
    \bottomrule
    \end{tabular}
    }
    \label{exp1}
    \vspace{-0.4cm}  
\end{table}
From Table~\ref{exp1}, we observe that the optimal selection of structural encoding typically varies across different datasets and base models. For example, RWSE tends to be more beneficial than LapPE for small molecular graph learning (\eg MolPCBA and ZINC), whereas the opposite is generally true on Peptides.
Notably, the performance gains are most pronounced when integrating PSE with Transformer, demonstrating its critical role in compensating for the absence of inherent structural sensitivity in Transformers. The last row shows the average improvement ($\%$) brought by GFSE on base models. The consistent improvements across different settings underscore the robustness and generalizability of GFSE, especially in contexts where traditional PSE fails to deliver. 

\subsection{Efficiency Evaluation}\label{appd:efficiency}
During the pre-training on eight datasets, the average time is around 30 to 40 minutes for each epoch with a single NVIDIA A40 48GB GPU. Total training time is less than two days, which is relatively efficient for a comprehensive multi-dataset pre-training process.

\begin{table}[htbp]\vspace{-0.4cm}
\centering 
\caption{Runntimes ($s$) of PSE computation on random synthetic graph with increasing numbers of nodes}\label{exp:efficiency}
\resizebox{0.35\textwidth}{!}{
\begin{tabular}{lcccc} \toprule
PSE / Graph size & 100 & 300 & 500 & 1000 \\ \midrule
LapPE & 2 & 9.25 & 34 & 155 \\
RWSE & 2 & 9.76 & 31.48 & 207 \\
\rowcolor{Gray}Pre-computation & 0.0007 & 0.001 & 0.003 & 0.006 \\
\rowcolor{Gray}GFSE Inference & 0.908 & 3.958 & 10.770 & 48.106 \\
\bottomrule
\end{tabular}}\vspace{-0.4cm}
\end{table} 
We compare the inference efficiency of GFSE with handcrafted positional encodings, such as LapPE and RWSE in Table~\ref{exp:efficiency} and Table~\ref{exp:efficiency2}. Specifically, we generate 1,000 synthetic Erdos-Rényi graphs for various graph sizes (100, 300, 500, and 1,000 nodes) and evaluate the time required for pre-computation and inference. 

\begin{table}[htbp]
\centering \vspace{-0.3cm}
\caption{Runntimes ($s$) on real-world graph dataset}\label{exp:efficiency2}
\resizebox{0.45\textwidth}{!}{
\begin{tabular}{lcccccc} \toprule
Dataset & ZINC & MolHIV & MolPCBA & Peptides & MNIST & CIFAR10 \\ \midrule
LapPE & 25 sec & 37 sec & 6.13 min & 73 sec & 96 sec & 2.55 min \\
RWSE & 11 sec & 58 sec & 8.33 min & - & - & - \\
\rowcolor{Gray}\textbf{GFSE} & 4.17 sec & 17.23 sec & 2.97 min & 15.21 sec & 49.38 sec & 1.27 min \\
\bottomrule
\end{tabular}}\vspace{-0.3cm}
\end{table}
As shown in the Table~\ref{exp:efficiency}, both LapPE and RWSE exhibit significant increases in computation time as the graph size grows. Pre-computation times required by GFSE inference remain minimal for all graph sizes, underlining the model's efficiency in this phase. In Table~\ref{exp:efficiency2}, we observe that GFSE demonstrates superior scalability in inference, making it a more efficient option for large-scale graph processing.


\end{document}